\documentclass{article}

\usepackage[preprint]{corl_2023} 

\title{On Hand-Held Grippers and the Morphological Gap in Human Manipulation Demonstration}

\author{
  Kiran Doshi
  \And
  Yijiang Huang
  \And
  Stelian Coros
  \AND
  \textnormal{Department of Computer Science} \\
  \textnormal{ETH Zürich} \\
  \footnotesize{\texttt{\{kiran.doshi, yijiang.huang, stelian.coros\}@inf.ethz.ch}}
}

\begin{document}
\maketitle


\begin{abstract}
    Collecting manipulation demonstrations with robotic hardware is tedious - and thus difficult to scale.
    Recording data on robot hardware ensures that it is in the appropriate format for Learning from Demonstrations (LfD) methods.
    By contrast, humans are proficient manipulators, and recording their actions would be easy to scale, 
    but it is challenging to use that data format with LfD methods.
    The question we explore is whether there is a method to collect data in a format that can be used with LfD while retaining some of the attractive features of recording human manipulation. 
    We propose equipping humans with hand-held, hand-actuated parallel grippers and a head-mounted camera to record demonstrations of manipulation tasks. 
    Using customised and reproducible grippers, we collect an initial dataset of common manipulation tasks.
    We show that there are tasks that, against our initial intuition, can be performed using parallel grippers.
    Qualitative insights are obtained regarding the impact of the difference in morphology on LfD by comparing the strategies used to complete tasks with human hands and grippers.
    Our data collection method bridges the gap between robot- and human-native manipulation demonstration. By making the design of our gripper prototype available, we hope to reduce other researchers effort to collect manipulation data.
\end{abstract}



\section{Introduction}
    \label{sec:introduction}
    
    Human manipulation capability is an ideal source of demonstration data to scale robotic manipulation skills using machine learning (ML). 
    Even though humans would be able demonstrators, the challenge in robotics is that, to date, it is expensive and tedious to collect demonstration data from humans in a robot-compatible format.
    There are two main approaches to record human demonstration data, which we categorise according to whether they are collected in a human-native or robot-native format.
    
    Real-world human demonstrations are available at scale on the internet (e.g. on YouTube), as well as in large datasets which have been curated, e.g.~\citep{grauman2022ego4d}. 
    While easy to collect and already readily available, the challenge is that, for now, leveraging this type of data in robotics remains difficult. There is a need for a mapping between human actions and those feasible for a robot~\citep{billard2020lfdsurvey} and, for certain tasks, human demonstration with their hands might not be informative for a robot with a parallel gripper. 
    
    Collecting data in a robot-native format, i.e., collected through tele-operation or kinesthetic teaching, enables robots to execute tasks it acquires by learning from human demonstration.  
    Recent work aims to make this type of data scale by being reusable across institutions \citep{walke2023bridgedata}.
    Nevertheless, it is unclear if this method is able to scale enough to enable similar breakthroughs as seen recently in natural language processing or computer vision.
    
    In this work, we propose a different method to collect human demonstration data in manipulation, which we believe would enable strong scaling while retaining a robot-compatible format.
    We equip the human collecting data with hand-held parallel grippers that are actuated by a pinching motion.
    Given that the grippers are wearable, and the objects are manipulated with those same grippers, they do not introduce any lag between perception and action, the action does not occur at a distance and the introduced mental overhead is minimal.
    The wearable grippers allow the human demonstrator to interact with the environment with the same interface as most robots do.
    Collecting demonstrations at scale with this approach should be possible in the same way as was done in \citep{grauman2022ego4d,goyal2017something}.
    It bridges the two modalities above by being simple for humans to use and does not introduce a large morphological gap (i.e. does not require a mapping between human and robot actions). 
    
    Furthermore, comparing a human at manipulation tasks wearing hand-held grippers versus their own hands, enables an analysis of the impact of the morphological gap on strategy and feasibility of manipulation tasks.
    By letting humans manipulate objects with the native robotic interface, we are able to base our analysis on the factorisation of the manipulation problem: 
    (i) General human intelligence allowing humans to form complex strategies to achieve their goals, and (ii) their capable, general-purpose, high degree of freedom (DOF) manipulator, the human hand. 

\begin{figure*}
        \centering
        \includegraphics[width=\textwidth]{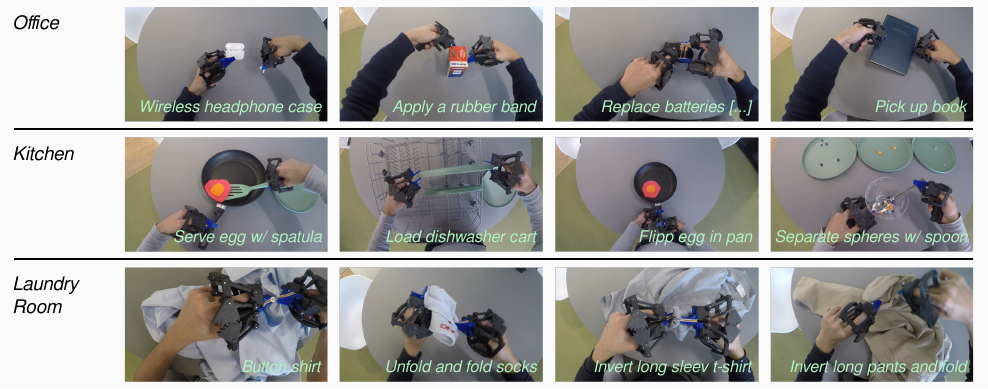}
        \caption{A selection tasks from the three environments \textit{Office}, \textit{Kitchen} and \textit{Laundry Room} we considered in our dataset. 
        }
        \label{fig:task_overview}
    \end{figure*}

    We discuss what we believe to be desirable in a data collection method, our prototype of the hand-held grippers, and how the data can be used to learn robotic skills.
    Using the prototype, we collected an initial dataset on a range of everyday human tasks, recording the observations using a head-mounted RGB camera.
    A selection of the recorded task demonstrations can be seen in \cref{fig:task_overview}.
    Using an analysis framework based on the above factorisation of the manipulation problem, we are able to show how the manipulation strategy changes for some interesting tasks such as \textit{Folding socks}, \textit{Inverting long pants} and \textit{Pick up book}. 
    Our aim with this work is to encourage other researchers to use this method to collect and share data, such that an additional source of manipulation demonstrations is available to the robotics community. 
    The proposed data collection method bridges the commonly used methods, and we believe this makes it useful in its own right while complementing the already existing data collection methods in robotic manipulation. 
    
\section{Hand-Held Parallel Grippers for Scalable Manipulation Demonstrations}
\label{sec:approach}

    \subsection{Necessary Features to Scale Manipulation Demonstration}
    \label{sec:features}
    Before considering our prototype, we set out the features we believe are necessary for a manipulation demonstration system.
    (i) The system is intuitive for a human operator to use without (much) training. 
    (ii) The system can be taken into any environment where a human is able to operate.
    (iii) The human demonstrator is able to perform dexterous manipulation tasks with it.
    (iv) The human demonstrator can demonstrate strategies which, when imitated by robotic hardware, can be analogously executed successfully. 

    
    \subsection{Hand-Held Parallel Gripper Prototype}
    Our hand-held gripper prototype is a light-weight, 3D printed parallel gripper. 
    It is held in the operators palm using the lower three fingers, while the thumb and the index finger are both placed in respective mounts for the gripper fingers. 
    The gripper can be opened and closed by a pinching motion of the thumb and the index finger.
    Details regarding manufacturing the prototype are in \cref{sec:app_grp_prototype}.
    
    Our prototype addresses the necessary features (see \cref{sec:features}) as follows.
    (i) The motion of the gripper when opening or closing corresponds to a pinch, which is the motion the operator uses to actuate it. 
    Additionally, the object is acted upon directly and not at a distance, which feels more natural to the operator compared with demonstrating in mid-air.
    Finally, the actuation mechanism is immediate and no lag is introduced.
    (ii) The only hardware required are the two wearable grippers in each hand and a head-mounted camera, which can be taken into most human environments.
    (iii) Dexterous tasks can be performed with our prototype (see \cref{sec:notable_tasks}).
    (iv) The hand-held grippers interface the environment in the same manner as the most common robotic manipulator, therefore we expect that the strategies demonstrated with the hand-held grippers can be performed by a bi-manual robot setup with parallel grippers. \\
    Additionally to addressing the necessary features, using the hand-held gripper prototype allows the demonstrator
    to build an intuition on how to tackle certain manipulation tasks. It allows them to experience the robot interface to the environment intuitively.
    
    
    \subsection{Analysis of Human Use of Hand-Held Parallel Grippers}
    \label{sec:analysis}

    With this analysis, we aim to understand which tasks that robots equipped with parallel grippers can be expected to perform.
    Our analysis is led by the factorisation of human manipulation performance introduced in \cref{sec:introduction}.
    The base assumption is that a person's computational performance, when using two parallel grippers in both hands, is on par with their performance when using their own hands.
    The term computational performance is used in the sense of humans' general intelligence employed in manipulation tasks.
    The following three cases structure our understanding.
    \begin{case}
        A human \textbf{can} solve a given task with a pair of hand-held parallel grippers using a similar strategy as when using their hands.
        \label{case:can_solve}
    \end{case}

    \begin{case}
        A human \textbf{cannot} solve a given task with a pair of hand-held parallel grippers (within a time limit) -- but can solve it using their hands.
        \label{case:cannot_solve}
    \end{case}

    \begin{case}
        A human needs to \textbf{significantly modify their strategy} to solve a given task with a pair of hand-held parallel grippers compared to when using their hands.
        \label{case:strategy_mod}
    \end{case}
    The implications of all tasks that fall under \cref{case:can_solve} are simple. Parallel grippers are not a limiting factor to completing the task successfully. Thus, if a robot cannot perform the same task, the bottleneck is due to the available computational methods (for sensing and decision making). 
    
    For tasks which relate to \cref{case:cannot_solve}, the discussion of the implications requires two sub-cases, \ref{case:cannot_solve}A and \ref{case:cannot_solve}B. 
    The conclusion for \cref{case:cannot_solve}A is that this is a task which cannot be solved with the parallel gripper and that a different gripper morphology is required to solve the task.
    In \cref{case:cannot_solve}B, our base assumption on equal computational performance does not hold and the human trying to solve the task with an end effector very different to his own is not able to come up with the correct strategy. 
    If this is the correct explanation, we can only conclude that humans cannot demonstrate how to solve the task, although a better learning algorithm (than the human mind) could find a solution. 
    
    \cref{case:strategy_mod} offers us interesting differences in strategies that arise due to the morphological difference. 
    It shows us that, for these tasks, naively imitating human demonstrations recorded using their hands is likely to fail. 
    This is because the difference in morphology causes such a strong change in the required strategy that the demonstration by hand would not be informative for the robot.
    If we can better understand the differences in strategies for various families of tasks and due to which morphological difference they are, this information can help inform computational methods which aim to bridge this gap.
    
    \subsection{Hand-Held Parallel Gripper Data for Robot Learning}
    With the data that we propose to collect, a central aim is that it can be used to learn a robot policy. 
    Although not implemented in this paper, we expect that policies can be learnt using LfD methods such as the state-of-the-art methods ACT \citep{zhao2023aloha} or Diffusion Policies \citep{chi2023diffusion}. 
    The dataset as we provide does not (yet) include position and orientation information of the grippers. 
    This information could be determined by equipping the grippers with a visual positioning system, e.g. as done in \citep{sanches2023HMI}. 
    The pose information serves as the end-effector targets for any 6+ DOF robotic manipulator, which can be tracked at the joint level by the robot by solving the inverse kinematics problem. 
    The objects manipulated can be represented by the recorded image alone, as it is feasible to learn a manipulation policy from image observations \citep{mandlekar2021matters,levine2016end}.


\section{Dataset and Analysis}
\label{sec:result}
	\subsection{Data Collection}
        We collected data for tasks which can be grouped in the following environments: \textit{Office},  \textit{Kitchen} and \textit{Laundry Room}. 
        The dataset that we collected to date was intended as a proof-of-concept and was also collected with the analysis questions of the last section in mind. 
        Therefore, all tasks are performed in an isolated setting outside of their usual environment.
        For analysis purposes, all tasks are recorded with both hand-held grippers and human hands.
        These tasks are an interesting sample of tasks that humans perform in their daily lives. 
        In future work, the dataset can be expanded to include demonstrations in varied environments.
        A selection of tasks for each environment can be found in \cref{fig:task_overview}. 
        The tasks for each environment and their classification according to the three cases (see \cref{sec:analysis}) are listed in \cref{tab:dataset_cases}.
        The classification into the cases is qualitative. 
        
        The following conclusions can be drawn from our analysis.
        Tasks which mainly require pick-and-place type actions fall under \cref{case:can_solve}. 
        In the tasks that fall under \cref{case:cannot_solve}, the pattern is that these tasks include tools that are adapted to the form factor of the human hand. 
        We found that many dexterous tasks are possible using parallel grippers, interestingly also tasks that we beforehand thought are unlikely to be possible with the grippers.
        Dexterous tasks often mainly fall under \cref{case:strategy_mod}, which takes the limited capabilities of the grippers versus the hands into account.
        In \cref{sec:limitations} we summarise which limitations of parallel grippers lead to tasks falling under \cref{case:cannot_solve,case:strategy_mod} and in \cref{sec:notable_tasks} we discuss the tasks with the most interesting features from our dataset.
        \newcolumntype{g}{>{\columncolor{Gray}}c}
        \newcolumntype{f}{>{\columncolor{Gray}}l}
        \begin{table}[ht]
            \centering
            \caption{All tasks according to environment and their classification into the Cases (C) in \cref{sec:analysis}.}
            \begin{tabular}{l f|g|g|g}
                \rowcolor{white}
                \textbf{Environment} & \textbf{Task} & \textbf{C1} & \textbf{C2} & \textbf{C3} \\
                \hline
                \rowcolor{white}
                \textit{Office} & Wireless headphone case & x & & \\
                & Applying a rubber band & & & x \\
                \rowcolor{white}
                & Opening a cardboard box & & & x \\
                & Pick up a book & & & x \\
                \rowcolor{white}
                & Apply paper clips & & & x\\
                & Plugging and unplugging cables & x & & \\
                \rowcolor{white}
                & Replacing batteries of a mouse & & & x \\
                & Use scissors & & x & \\
                \rowcolor{white}
                & Open and close resealable bag & x & & \\
                \hline
                \hline
                \rowcolor{white}
                \textit{Kitchen} & Flip egg in pan & & x & \\
                & Serve (art.) egg with rubber spatula & x & & \\
                \rowcolor{white}
                & Serve (art.) egg with wooden spatula & x & & \\
                & Separate spheres on plate with spatula & x & & \\
                \rowcolor{white}
                & Separate spheres on plate with spoon & x & & \\
                & Separate spheres in bowl 1 with spoon & & & x \\
                \rowcolor{white}
                & Separate spheres in bowl 2 with spoon & & & x \\
                & Load dishwasher cart & & & x \\
                \rowcolor{white}
                & Replace trash bag & x & & \\
                & Manipulate cardboard box & & & x \\
                \hline
                \hline
                \rowcolor{white}
                \textit{Laundry Room} & Unfold and fold socks & & & x\\
                & Invert long sleeve t-shirt & & & x \\
                \rowcolor{white}
                & Invert short sleeve t-shirt & x & & \\
                & Close zip & & & x \\
                \rowcolor{white}
                & Fold t-shirt & x & & \\
                & Invert long pants and fold & & & x\\
                \rowcolor{white}
                & Zip and button pants & & & x \\
                & Button shirt & & x & x \\
                \rowcolor{white}
                & Invert short pants and fold & & & x\\
            \end{tabular}
            \label{tab:dataset_cases}
        \end{table}
                
        \subsection{Observed Limitations in Manipulation with Hand-Held Parallel Grippers}
        \label{sec:limitations}
        Tasks that fall under Cases 2 and 3 (see Section \ref{sec:analysis}) allow us to analyse the ways in which the parallel gripper is limited compared to the human hand.
        This helps us to understand if human hand demonstrations could be transferred to a robot. In this section, we summarise our conclusions based on the performed tasks. 
	
        \begin{limitation}[More DOFs on Hand]
            \label{lim:more_dof}
            The human hand has more DOFs than the parallel gripper. While for most manipulation tasks by hand not all are used, especially the additional DOFs in the thumb and index finger often provide a large advantage over the gripper.
        \end{limitation}
        
        \begin{limitation}[Larger Gripping Width]
            \label{lim:grip_width}
            In the extreme case, the functional gripping width of the human hand spans from the thumb to the ring finger, which is significantly more than what is possible with most parallel grippers. This enables more simple pick-and-place strategies than with a parallel gripper.
        \end{limitation}
        
        \begin{limitation}[More Points of Contact]
            \label{lim:poc}
            The human hand has five digits, which can all be used to provide points of contact and apply force when needed. With the gripper, there is either one point of contact (or two if the gripper is used in a non-prehensile way). The larger number of points of contact on the human hand is especially useful for tasks where an opening needs to be created.
        \end{limitation}
        
        \begin{limitation}[Not Able to Reach into Openings]
            \label{lim:reach_opening}
            Compared to the human hand, reaching into small openings is often impeded by the size of the gripper design.
            Additionally, when the gripper is inside an object which occludes it, is is not possible to determine what is being manipulated.
            In contrast, using tactile feedback it is always possible for the human to approximately know what they are manipulating.
        \end{limitation}

        \subsection{Notable Tasks}
        \label{sec:notable_tasks}
        For \cref{case:cannot_solve,case:strategy_mod} we provide a description of the most notable and interesting tasks. 
        We aim to furnish the reader with a visual intuition of the experience of providing demonstrations with the hand held grippers and how it compares to solving the same tasks with their hands. 
        The timelines in all figures are relative to the particular task displayed and are approximate. 
        For every figure we provide a video version which can be found here\footnote{\url{https://youtube.com/playlist?list=PLvom4vstj7Pc5HGb1E1f0uO220QGQIgqc&si=jNN_J7Z4qUvj2LPC}}.
        More notable tasks can also be found in \cref{sec:futher_notable_tasks}.
        
\begin{figure*}
            \centering
            \includegraphics[width=\textwidth]{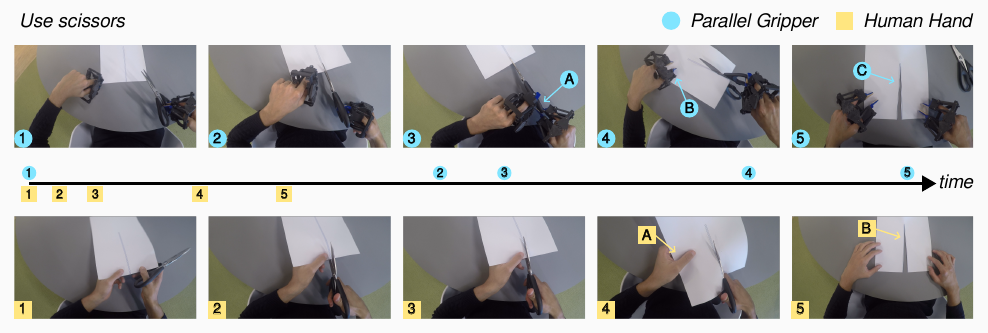}
            \caption{
                \textcolor{mybluedark}{(A)} Both grippers needed to actuate the scissors, paper cannot be held while cutting. 
                \textcolor{mybluedark}{(B)} Left gripper feeds paper into partially opened scissors.
                \textcolor{mybluedark}{(C)} Cut is not very precise.
                \textcolor{myyellowdark}{[A]} Paper and scissors are aligned precisely, the left hand can feed the paper while the right hand makes a cut. 
                \textcolor{myyellowdark}{[B]} Cut is very precise. 
            }
            \label{fig:use_scissors}
        \end{figure*}

        \paragraph{\cref{case:cannot_solve}}
        The task \textit{Use scissors} cannot be solved with two parallel grippers, mainly because it requires two grippers to hold and actuate the scissors, while a third gripper would be required to hold the paper being cut (\cref{lim:poc}). While it is possible to cut the paper (see \cref{fig:use_scissors}) using the grippers, because there is little control, the precision is significantly reduced.

\begin{figure*}
            \centering
            \includegraphics[width=\textwidth]{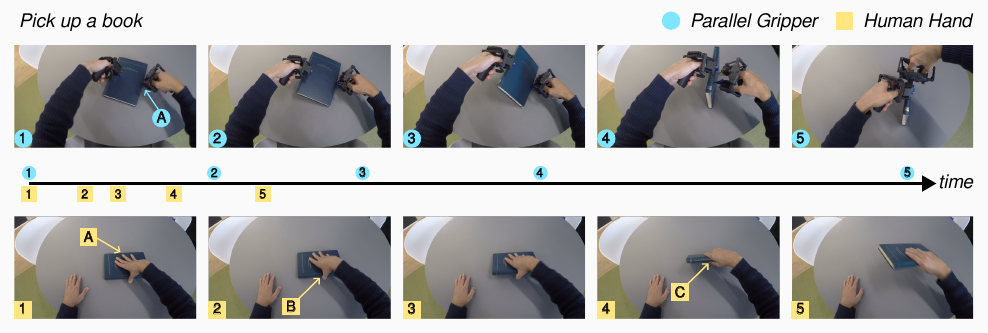}
            \caption{
                \textcolor{mybluedark}{(A)} Right gripper provides counter force to prevent the book from sliding. 
                \textcolor{myyellowdark}{[A]} Fingers on the book cover provide counter force to prevent the book from sliding. 
                \textcolor{myyellowdark}{[B]} Additional DOF in thumb used to lever book from table. 
                \textcolor{myyellowdark}{[C]} Fingers and thumb reposition in anticipation of grip.}
            \label{fig:pick_up_book}
        \end{figure*}

        \paragraph{\cref{case:strategy_mod}}      
        \label{sec:results_strategy_mod}
        In \textit{Pick up a book}, two grippers are required versus only one hand (see \cref{fig:pick_up_book}). 
        With a human hand, the task can be performed with one hand using the fingers on the cover to provide a counter force, while the additional DOFs of the thumb are used to lever the book from the table.  
        When using grippers, one gripper is needed to provide the counter force for the pivot, while the other gripper lifts the book until it is possible to grip the short edge after a reorientation.
        
\begin{figure*}
            \centering
            \includegraphics[width=\textwidth]{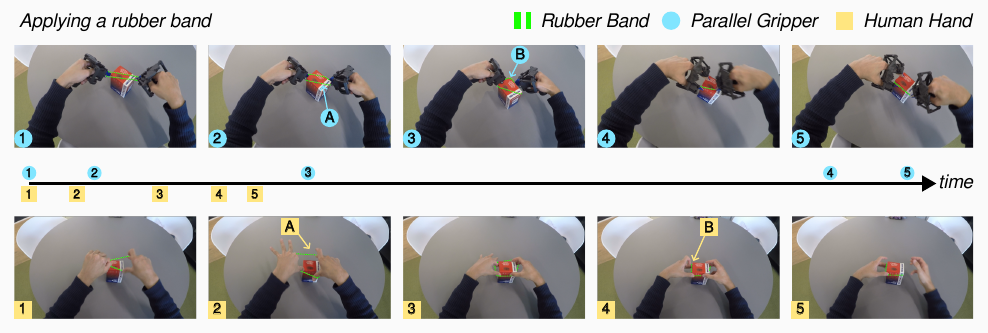}
            \caption{
                \textcolor{mybluedark}{(A)} The band is stretched along the diagonal of the box. 
                \textcolor{mybluedark}{(B)} The band is pulled to the corner of the box with both grippers, while ensuring that the rubber band stays in contact with the box over the diagonal. 
                \textcolor{myyellowdark}{[A]} The rubber band is stretched over the box using four fingers. 
                \textcolor{myyellowdark}{[B]} Fingers successively release the band onto the box. 
                }
            \label{fig:rubber_band}
        \end{figure*}

        In the case of \textit{Applying a rubber band} (see \cref{fig:rubber_band}), manipulation can be done with either one or two hands, where the fact that the hand has multiple fingers and thus contact points to stretch the band is used. When using the gripper by picking the band, only two contact points are available (\cref{lim:poc}). Using this approach, the elastic band cannot be stretched open. The friction between the band and the surface is used to keep the band stable while it is spanned over only two corners. Once spanned over three corners, the band can be pulled over the last with one gripper while the other stabilises the entire object.
        
\begin{figure*}
            \centering
            \includegraphics[width=\textwidth]{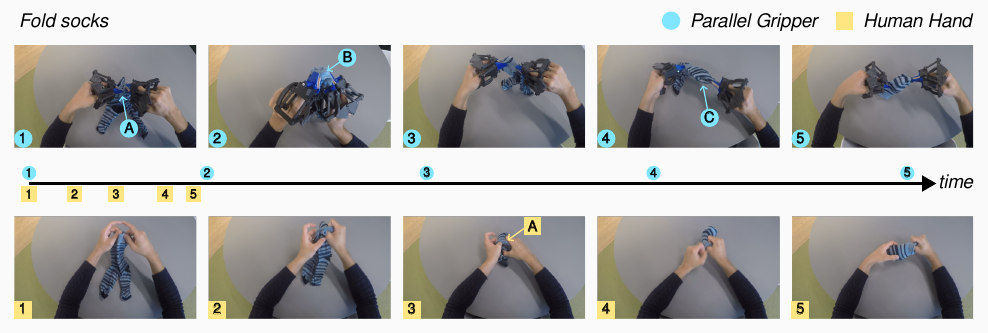}
            \caption{
                \textcolor{mybluedark}{(A)} Both socks are aligned and the left gripper grips both sock bands. 
                \textcolor{mybluedark}{(B)} Rubber band of left sock is inverted over the right sock.
                \textcolor{mybluedark}{(C)} To make space for the sock to be pushed into, the right gripper pulls on the outside.
                \textcolor{myyellowdark}{[A]} In one move, the entire sock band is inverted and the sock is stuffed into the inverted sock. 
                }
            \label{fig:fold_socks}
        \end{figure*}

        To \textit{Fold socks}, the change of strategy needs to account for 
        the \cref{lim:poc,lim:reach_opening}.
        With our hands, we are able to fold the socks with both hands in one continuous motion.
        With the gripper this is not possible, the approach demonstrated requires many re-grasping steps and is iterative (see \cref{fig:fold_socks}).
        The key to folding socks with the gripper is to invert the elastic band (at the top of the sock) of one sock over the other.
        Once the elastic is inverted, using a mixture of stuffing on one side and pulling the inverted side outward on the other, the folding is completed. 
        This is a slow process and significantly slower than what is possible with the human hand. \\
        This task is a good example of how the researcher can develop their intuition by using the parallel grippers. 
        We assumed that it would not be possible to fold socks without human hands, 
        as the common folding strategy requires all features of the hand that the grippers are missing.
        By using the hand-held parallel grippers, we discovered that a change of strategy makes the task possible.
        
\begin{figure*}
            \centering
            \includegraphics[width=\textwidth]{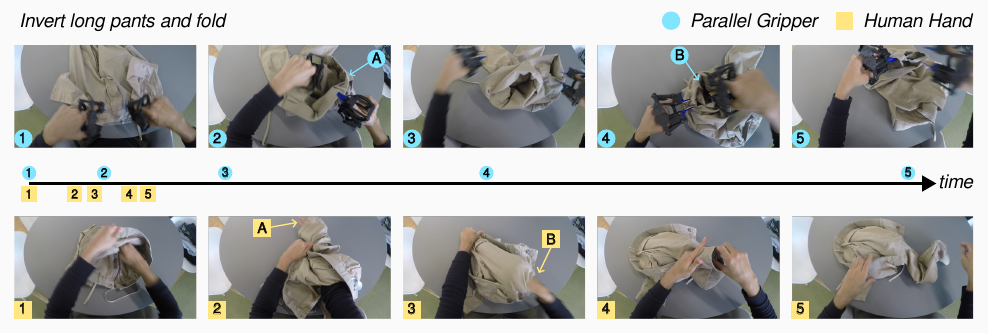}
            \caption{
                \textcolor{mybluedark}{(A)} The waistband is inverted to a stable position. \textcolor{mybluedark}{(B)}  While the left gripper fixes the pants, the right gripper pulls the leg from the outside, iteratively inverting the leg.
                \textcolor{myyellowdark}{[A]} The hand reaches into the pant leg and looks out at the end of the leg \textcolor{myyellowdark}{[B]} Gripping the hem, the pant leg is inverted.
                }
            \label{fig:invert_long_pants}
        \end{figure*}

        The main limitation for \textit{Invert long pants and fold} is that, depending on the cross-section of the pant leg, it is not possible to reach into the leg with the hand-held grippers (\cref{lim:reach_opening}). Therefore, to invert the pant leg, the strategy is to pull numerous times from the outside and thereby slowly pull out the pant leg until it is fully inverted (see \cref{fig:invert_long_pants}). 
\section{Related Work}
\label{sec:related_work}

Our proposed method for collecting demonstration data bridges the available data modalities and their collection methods.
At one end of the spectrum, data is recorded with robot hardware, where the human demonstrator interacts with the robotic system either by tele-operation or kinesthetic teaching.
BridgeData V2 \citep{walke2023bridgedata} is collected by tele-operation with a VR controller in multiple similar lab environments.  
RH20T \citep{fang2023rh20t} is collected by tele-operation, although in this case with a haptic input device for the human operator. 
The \textsc{RoboTurk} \citep{mandlekar2018roboturk} approach crowd-sources data collection, by allowing people to remotely control a robot arm and complete tasks using a tele-operation interface controlled with an iPhone application.
These large datasets (considering the standards of robotics) need to be recorded in a lab setting, while our grippers can be taken outside of the lab, thus enabling more diverse data content.

At the other end of the spectrum, humans are recorded performing manipulation tasks in their natural environments, capturing human manipulation diversity. 
Examples of such datasets are \textsc{Ego4D} \citep{grauman2022ego4d} and \textsc{Something Something} \citep{goyal2017something}.
\textsc{Ego4D} is not primarily a robotic dataset. 
However, given that it captures a lot of human manipulation, it can become a source for robotic LfD, if the methods to transfer such demonstrations to a robot embodiment are mature enough.
Nevertheless, in this paper, we have shown that there are tasks in which the strategy strongly differs between human hand and parallel gripper.
For these tasks, it is likely that demonstrations need to include parallel grippers.
Our method can therefore complement human data collection initiatives with data that can be used for robotics.

Prior work has also focused on trying to ease the difficulty of collecting robotic data. 
\citep{zhao2023aloha,wu2023gello} propose open-source setups where the tele-operation input device respects the kinematic constraints of the robot arm and is controlled at the joint level, reducing latency and providing a more intuitive tele-operation method.
\citep{song2020grasping,young2021visual} propose using a grasping device to record demonstrations.
The device is limited compared to our prototype and thus restricts demonstrations to pick and place tasks and does not include more complex, dexterous tasks.
In \citep{iyengar2018dataset}, the robotic gripper hardware is used to record grasp demonstrations and the demonstration system is such that it is restricted to uni-manual grasps. 
The work in \citep{sanches2023HMI} is closely related to our approach.
The hand-held gripper prototype differs from ours in that it is larger, as they mount the actual robotic gripper to the human demonstrator by strapping it to the wrist.
The gripper is actuated using a button and their setup allows the option of mounting different end-effectors.
Our work is more focused on showing the potential of how our gripper can be used to scale data collection in LfD and the comparison of hand and gripper strategies.




\section{Conclusion}
\label{sec:conclusion}

    In this work, we proposed a data collection method that can help scale demonstration data collection for robot manipulation. 
    Collecting data with hand-held parallel grippers complements and bridges current methods of collecting data.
    A wide range of dexterous tasks are demonstrated in the dataset, with tasks being demonstrated by the same human subject using the grippers and their own hands.
    This enables a systematic analysis of the intricate interplay between gripper morphology and manipulation strategy.
    It also shows interesting modifications in strategy required to compensate for the morphological difference of the gripper.
    We leave it to future work to use the data to learn a manipulation policy.
    We hope that this paper provides convincing initial results to invite the robot learning community to join forces with us to increase the size of the hand-held gripper dataset in varied environments.


\clearpage
\acknowledgments{The authors would like to thank Oliver Stark for the mechanical design of the hand-held parallel grippers and Miguel Zamora for fruitful discussions. YH acknowledges the generous support provided by the ETH Zurich Postdoctoral Fellowship programme.}


\bibliography{example}  

\newpage
\appendix
\section{Hand-Held Gripper Prototype}
    \label{sec:app_grp_prototype}
    The prototype parts are printed with Onyx (by Markforged).
    Certain parts which require more rigidity are additionally reinforced with carbon fibre. 
    The main structure of the gripper fingers is printed with PLA+ and a pad of a softer material, TPU 60A, to increase grip.
    The TPU pad is printed on the surface that is in contact with the manipulated objects.
    In addition to the 3D printed parts, ball bearings are used to ensure friction-free motion in the actuation mechanism.
    We aim to open-source the design of our hand-held gripper prototype soon.
\section{Further Notable Tasks}
    \label{sec:futher_notable_tasks}
    \subsection{\cref{case:cannot_solve}}
\begin{figure*}
            \centering
            \includegraphics[width=\textwidth]{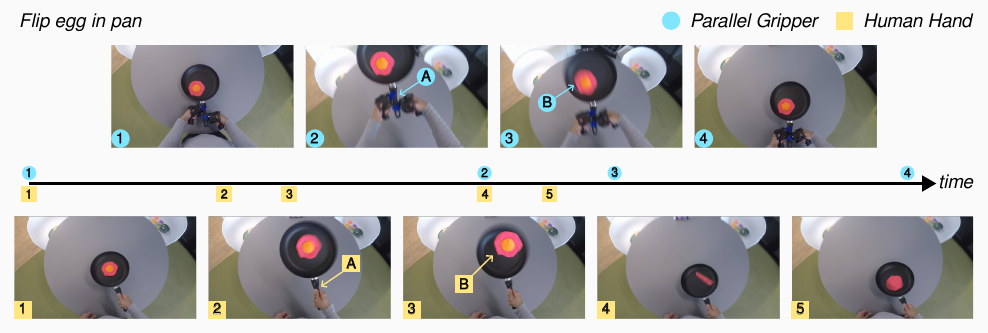}
            \caption{
                \textcolor{mybluedark}{(A)} The pan handle is held with two grippers, but the grip is not good enough to allow for quick deceleration. 
                \textcolor{mybluedark}{(B)} Deceleration is too slow such that friction between the egg and pan is not overcome. 
                \textcolor{myyellowdark}{[A]} Firm grip between pan and handle, allows close-to-instant deceleration.
                \textcolor{myyellowdark}{[B]} Friction force between egg and pan is overcome, the egg leaves the pan surface. 
                }
            \label{fig:flip_egg_in_pan}
        \end{figure*}

        \textit{Flip egg in pan} is difficult to perform with the hand-held grippers due to the fact that the grip of the pan handle with the grippers is not as firm as with the human hand (see \cref{fig:flip_egg_in_pan}). 
        The grip not being as firm is due to \cref{lim:poc,lim:grip_width}. 
        The larger gripping width and the increased number of contact points better allow force to be transferred from the hand to the pan handle. 
        This force transfer enables rapid deceleration which causes the egg to leave the surface of the pan.
        
    \subsection{\cref{case:strategy_mod}}
\begin{figure*}
            \centering
            \includegraphics[width=\textwidth]{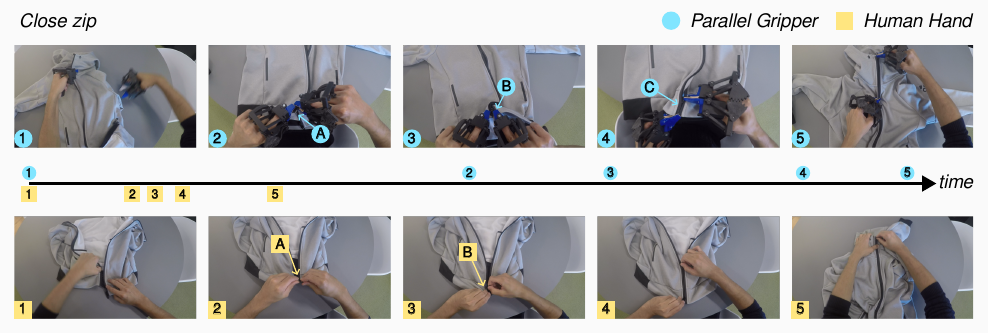}
            \caption{
                \textcolor{mybluedark}{(A)} The insertion pin is exposed. 
                \textcolor{mybluedark}{(B)} Using both grippers, the pin is manoeuvred into the slider.
                \textcolor{mybluedark}{(C)} Grippers repositioned, left gripper holds retainer box while right gripper pulls on slider.
                \textcolor{myyellowdark}{[A]} Only one hand needed to expose insertion pin, slider can already be gripped. 
                \textcolor{myyellowdark}{[B]} One finger repositioned to hold retainer box. 
            }
            \label{fig:close_zip}
        \end{figure*}

         The \textit{Close zip} task, only requires slight changes compared to a human doing it with their hands (see \cref{fig:close_zip}).
        The main difference, due to \cref{lim:poc,lim:more_dof}, is that with the grippers regrasping steps (i.e., letting go of the gripped part of the object) are required while fixing the insertion pin in the retainer box. With the human hand, the free fingers can be employed in changing the grasp without letting go of the object.
        
\begin{figure*}
            \centering
            \includegraphics[width=\textwidth]{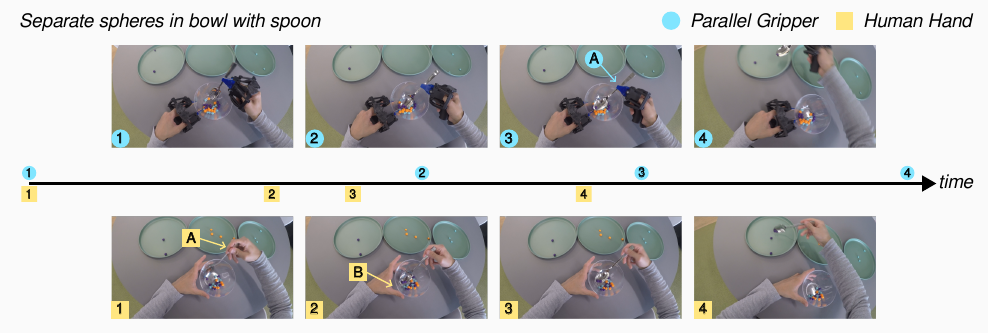}
            \caption{
                \textcolor{mybluedark}{(A)} Degree of manoeuvrability with a spoon is not very large because the contact is fixed.
                \textcolor{myyellowdark}{[A]} Orientation of spoon can be modified in place.
                \textcolor{myyellowdark}{[B]} Orientation of bowl can be modified to increase the ease of manoeuvrability.  
            }
            \label{fig:separate_spheres}
        \end{figure*}

        In \textit{Separate spheres in bowl with spoon} (see \cref{fig:separate_spheres}) the strategy changes in two ways, both due to \cref{lim:more_dof} and both of which significantly slow down the speed at which the task can be completed. The first is that with the gripper re-grasping is necessary to change the orientation of the spoon in the gripper. The second is that the free gripper that does not hold the spoon cannot be used effectively to modify the orientation of the bowl to simplify the manipulation with the spoon. Using the hands, it is very easy to reorient the bowl.
    
\end{document}